\documentclass[review]{elsarticle}

\usepackage{framed,multirow}
\usepackage{hyperref}
\usepackage{times}
\usepackage{epsfig}
\usepackage{graphicx}
\usepackage{amsmath}
\usepackage{amssymb}
\usepackage{setspace}
\usepackage{url}
\usepackage{xcolor}
\usepackage{multirow}
\usepackage{amsmath}
\usepackage{array}
\usepackage{geometry}
\usepackage{adjustbox}
\usepackage{caption,booktabs}
\usepackage{amssymb}
\usepackage{latexsym}

\usepackage[ruled, lined, commentsnumbered, longend]{algorithm2e}
\usepackage{xcolor}

\makeatletter
\def\ps@pprintTitle{%
 \let\@oddhead\@empty
 \let\@evenhead\@empty
 \def\@oddfoot{}%
 \let\@evenfoot\@oddfoot}
\makeatother

\begin{document}

\begin{frontmatter}

\title{LLEDA - Lifelong Self-Supervised Domain Adaptation}

\author[address1]{Mamatha Thota}

\author[address2]{Dewei Yi}

\author[address2]{Georgios Leontidis\corref{cor}}
\cortext[cor]{Corresponding author}
\ead{georgios.leontidis@abdn.ac.uk}

\address[address1]{School of Computer Science, University of Lincoln, LN6 7TS, Lincoln, United Kingdom}
\address[address2]{School of Natural and Computing Sciences \& Interdisciplinary Centre for Data and AI, University of Aberdeen, AB24 3UE, Aberdeen, United Kingdom}

\begin{abstract}
Humans and animals have the ability to continuously learn new information over their lifetime without losing previously acquired knowledge. However, artificial neural networks struggle with this due to new information conflicting with old knowledge, resulting in catastrophic forgetting. The complementary learning systems (CLS) theory \cite{mcclelland1995there, kumaran2016learning} suggests that the interplay between hippocampus and neocortex systems enables long-term and efficient learning in the mammalian brain, with memory replay facilitating the interaction between these two systems to reduce forgetting. The proposed Lifelong Self-Supervised Domain Adaptation (LLEDA) framework draws inspiration from the CLS theory and mimics the interaction between two networks: a DA network inspired by the hippocampus that quickly adjusts to changes in data distribution and an SSL network inspired by the neocortex that gradually learns domain-agnostic general representations. LLEDA’s latent replay technique facilitates communication between these two networks by reactivating and replaying the past memory latent representations to stabilise long-term generalization and retention without interfering with the previously learned information. Extensive experiments demonstrate that the proposed method outperforms several other methods resulting in a long-term adaptation while being less prone to catastrophic forgetting when transferred to new domains.
\end{abstract}

\end{frontmatter}


\section{Introduction}

Deep neural networks have shown near human-level capabilities in many fundamental computer vision tasks \cite{krizhevsky2017imagenet, he2021interpretable, Gong2021, ribeiro2022learning, ren2022graph}. Humans and animals can continuously acquire new information over their lifetime without catastrophically forgetting the prior knowledge learned. This ability to continually learn over time by accommodating new knowledge while retaining the previously learned knowledge is referred to as lifelong or continual learning (in our paper, we will continue to refer to it as lifelong learning). However, artificial neural networks lack these capabilities as new information interferes with previously learned knowledge and sometimes the old knowledge completely gets overwritten by the new one, leading to impaired performance \cite{mccloskey1989catastrophic}. The root cause of catastrophic forgetting is that learning necessitates changes in the weights of a neural network, however, these changes also result in the forgetting of previous learning.

The focus of this paper is on lifelong domain adaptation, in which the model is trained on multiple sequential domains, continuously adapting to new domains with changing distributions as they become available, while maintaining its knowledge of previously encountered domains.

Domain adaptation (DA) methods based on deep learning have received significant attention in recent years for mitigating the domain shift from the training domain to the inference domain \cite {long2015learning, thota2020multi, ganin2016domain, Thota_2021_CVPR}, and have even been suggested as transformative technologies in settings such as agriculture \cite{durrant2021might, onoufriou2023premonition} and arts \cite{pasqualino2021unsupervised}. However, current domain adaptation methods operate under the assumption that datasets from both the source and the target domains are accessible at the same time during training, which may not be feasible in practice. In addition, DA algorithms require fully labeled datasets, even state-of-the-art Unsupervised Domain Adaptation (UDA) methods need access at least to the source labeled dataset. Therefore, these algorithms require persistent manual annotation, which is time-consuming, cumbersome and expensive. Finally, just updating the underlying model will not be sufficient, as the model would likely forget the past learned domain information resulting in catastrophic forgetting. Acknowledging these issues, we propose LLEDA that addresses both catastrophic forgetting and domain-agnostic knowledge transfer using solely unlabeled datasets with access to a single domain at any given time.

The mammalian brain can continually acquire, process, consolidate, retrieve, and infer knowledge over time without catastrophically forgetting the previously learned information which can be explained using CLS theory \cite{mcclelland1995there, kumaran2016learning}. It suggests that efficient learning in the mammalian brain requires two learning systems: the neocortex and the hippocampus. The first system gradually acquires structured generalised knowledge, while the second system quickly learns the specific experiences, and the interplay between these two systems enables long-term retention. It also implies that memory replay is the mechanism that facilitates interaction between these two systems to consolidate and stabilise new memories for long-term generalization to reduce catastrophic forgetting.

Recently, a study by \cite{hayes2021replay} identified that the existing lifelong learning techniques are missing few biological elements. They highlight that many existing approaches solely focus on modelling the cortex directly and do not have a rapid learning network which is essential for facilitating effective lifelong learning in the brain. Additionally, the study also points out that none of the current methods employs information from the neocortex-inspired network to influence the training of the hippocampal-inspired network, whereas, in biological networks, the neocortex influences learning in the hippocampus and vice versa. 

Our proposed LLEDA network attempts to solve the first issue by utilizing two distinct networks, DA network for rapid learning and the SSL network for gradual acquisition. LLEDA mimics the interplay between the neocortex and the hippocampus, where the hippocampal-inspired DA network functions as a rapid acquisition mechanism to adapt the distribution shift between the given data stream and the data from memory, and the neocortex-inspired SSL network works like a gradual learning mechanism to generalise the representations by gradually acquiring structured knowledge using self-supervised techniques enabling effective lifelong learning.  LLEDA’s Latent memory replay facilitates communication between these two networks by reactivating the neural activity patterns representing previous experiences to stabilise new memories for long-term generalization and retention without interfering with the previously learned information. LLEDA attempts to address the second issue by querying the information from the neocortex to influence the training of the hippocampal-inspired network during training. 

Overall, our framework reduces catastrophic forgetting, while facilitating domain-agnostic knowledge transfer without accessing labeled data both from the source and target domains at any given time. To the best of our knowledge, this is an area of domain adaptation that has not yet been explored. In summary, our work makes the following contributions:
\begin{enumerate}
    \item Inspired by the CLS theory, LLEDA mimics the interplay between the DA network which helps to rapidly adapt the distribution shifts between domains, and the SSL network that helps with the gradual acquisition of domain-agnostic general representations, and the latent representations replay technique helps to replay the past memory representations, instead of raw image pixels to overcome catastrophic forgetting. 
    \item Our proposed self-supervised based approach does not require access to either source or target labels, hence saving time and effort to annotate data and assisting with the labeling bias.
    \item Extensive empirical results demonstrate that our method performs competitively across several benchmarks, when compared against other approaches.
\end{enumerate}

The rest of the paper is organized into several sections. Section-\ref{rw} offers an extensive literature review on Domain Adaptation (DA), Self-Supervised Learning (SSL), and Continual Learning (CL). In the section-\ref{method}, we present a detailed explanation of our LLEDA framework, which comprises three main components: Generalized Feature Learning, Domain-Specific Representation Learning, and Latent Replay. The section-\ref{exp} discusses the datasets used, training methodology, implementation details, and presents results, analysis, and ablation studies. Finally, section-\ref{con-fw} provides a summary of the paper's findings and outlines potential directions for future research.

\begin{figure*}[!t]
\centering
\hspace*{-1mm}
\includegraphics[width=0.99\textwidth]{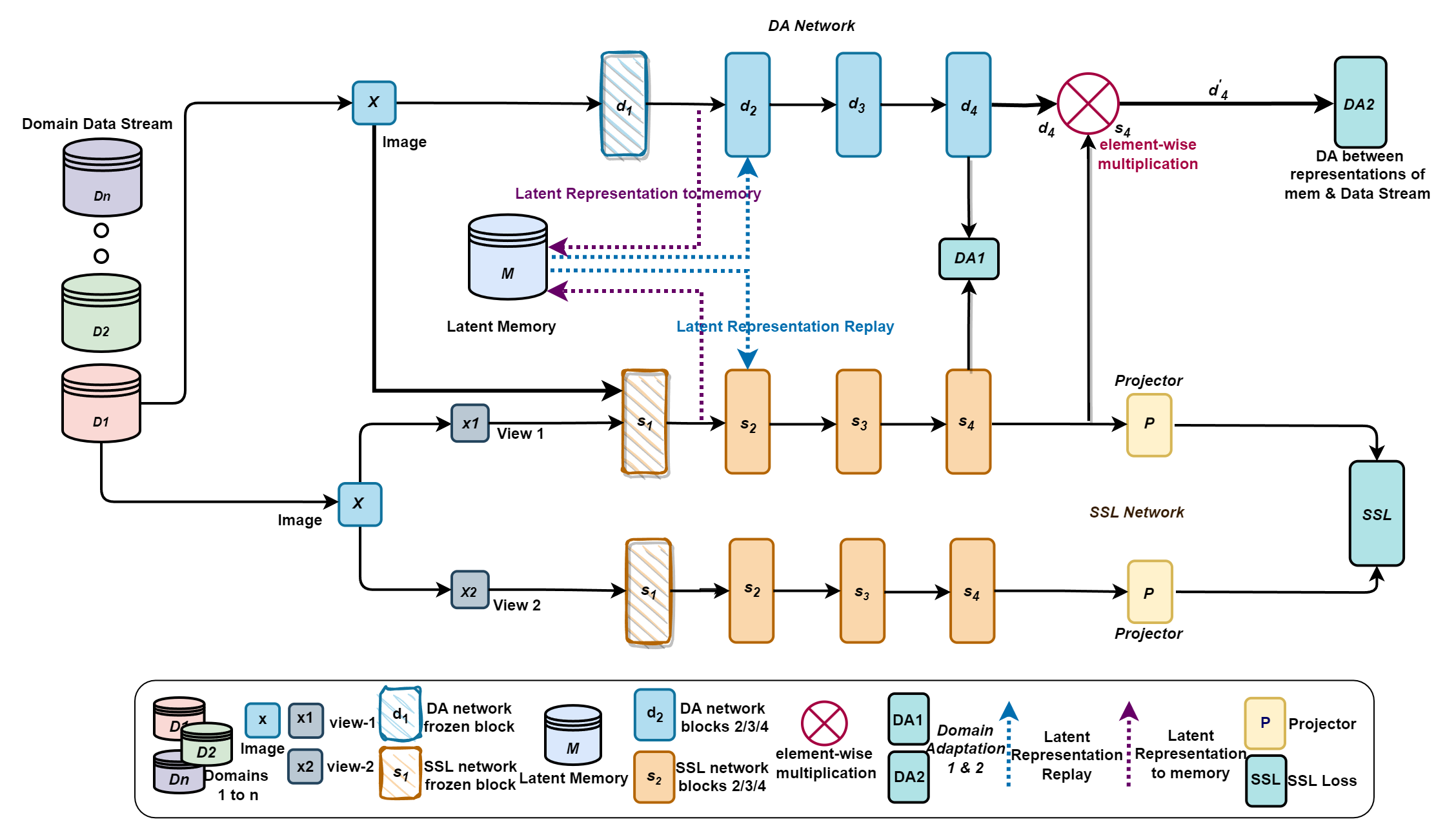}
\vspace*{-0.1cm}
\caption{Overview of the proposed LLEDA architecture. LLEDA consists of rapid learning \textbf{DA network} and gradual learning \textbf{SSL network}. The \textbf{SSL network} learns generic representations using self-supervised learning and \textbf{DA network} helps to overcome domain shift by optimising DA loss at two levels, \textbf{DA1}- MMD loss between the representations of $d_4$ and $s_4$, and \textbf{DA2} - MMD loss between memory representations and current data representations.}
\label{fig-network}
\end{figure*}

\section{Related Work} \label{rw}

\textit{Domain Adaptation}: Under the assumption of independent and identically distributed (iid) data, a deep neural network trained on one set of data is expected to perform well on a new, unseen set of data. However, this assumption may not always hold in real-world applications due to the discrepancy between domain distributions, and applying the trained model to the new dataset may also result in negative performance. Domain adaptation is a special case of transfer learning where the goal is to learn a discriminative model in the presence of domain shift between source and target datasets. Various methods have been introduced to minimise the domain discrepancy in order to learn domain-invariant features. Some involve adversarial methods like DANN \cite{ganin2016domain}, ADDA \cite{tzeng2017adversarial} that help align source and target distributions. Other methods propose aligning distributions through minimising divergence using popular methods like maximum mean discrepancy \cite{gretton2012kernel, long2015learning, long2017deep, jiang2020eeg, Gong2021, thota2020multi, Thota_2021_CVPR, zhang2022multi}, correlation alignment \cite{sun2016deep, chen2019joint, patel2022cross}, and the Wasserstein metric \cite{chen2019deep, lee2019sliced}. MMD was first introduced for the two-sample tests of the hypothesis that two distributions are equally based on observed samples from the two distributions \cite{gretton2012kernel}, and this is currently the most widely used metric to measure the distance between two feature distributions. The Deep Domain Confusion Network proposed by Tzeng et al.\cite{tzeng2014deep} learns both semantically meaningful and domain invariant representations, while  Long et al. proposed DAN \cite{long2015learning} and JAN \cite{long2017deep} which both perform domain matching via multi-kernel MMD (MK-MMD) or a joint MMD (J-MMD) criteria in multiple domain-specific layers across domains. 

\textit{Self-Supervised Learning}: Self-Supervised Learning (SSL) is a paradigm developed to learn visual features from unlabeled data. Recently, numerous SSL approaches have shown significant performance sometimes even surpassing, the performance of supervised baselines \cite{chen2020simple, chen2020big, he2020momentum, durrant2022hyperspherically, bardes2021vicreg, zbontar2021barlow, grill2020bootstrap, chen2021exploring, ma2023two, manova2023s, alkhalefi2023semantic, durrant2023hmsn}. These methods use image augmentation techniques to generate multiple views of a given image and learn a model that is invariant to these augmentations. Most recent approaches are divided into two main categories, contrastive and non-contrastive methods. Contrastive methods learn an embedding space where positive pairs are pulled together, whilst negative pairs are pushed away from each other \cite{chen2020simple, chen2020big, he2020momentum}. Non-contrastive methods on the other hand remove the need for explicit negative pairs either by using distillation or by regularization of the variance and covariance of the embeddings \cite{grill2020bootstrap, chen2021exploring, bardes2021vicreg, zbontar2021barlow}. However, none of these works studied the ability of SSL methods to learn continually and adaptively if they are applied directly. Moreover, very few works have attempted to use SSL in the lifelong domain adaptation setting, e.g. \cite{tang2021gradient} is designed using contrastive learning, so it lacks the capability to adapt using other SSL paradigms. \cite{SCHUTERA2021104079} trains model stepwise by generating pseudo labels and fine-tuning on intermediate domains until it reaches the target domain, this model can adapt well only if the domain shift is small between the intermediate domains, and it also uses source-labeled data. In this paper, we present a general-purpose framework to incorporate self-supervised learning approaches into the lifelong learning process to extract generalised representations.

\textit{Continual learning}:
Continual learning strategies aim to find the right balance between preventing catastrophic forgetting and acquiring new information. According to \cite{DBLP:journals/corr/abs-1802-07569}, catastrophic forgetting can be  mitigated using model regularization, memory replay or by expanding and training the network. Regularization methods identify the network weights that contribute significantly to retaining knowledge about a previously learned task and then consolidate them when the model is updated to learn the subsequent tasks \cite{DBLP:journals/corr/LiH16e, DBLP:journals/corr/KirkpatrickPRVD16, DBLP:journals/corr/JungJJK16}. On the other hand, dynamic architectures modify the model’s underlying architecture by dynamically accommodating neural resources as it learns new patterns \cite{DBLP:journals/corr/RebuffiKL16, DBLP:journals/corr/abs-1708-01547, 726791}. Alternatively, the model can be expanded progressively to learn the new tasks using added weights that propose ways of constraining the tasks’ objectives to avoid forgetting \cite{DBLP:journals/corr/abs-1711-10563, DBLP:journals/corr/Lopez-PazR17, DBLP:journals/corr/abs-1810-11910}. CLS and replay methods rely on memory replay by storing samples from old distributions and regularly feeding them back to the model to overcome catastrophic forgetting. Some of the existing CL methods \cite {chaudhry2019tiny, rebuffi2017icarl, DBLP:journals/corr/abs-2110-00175} store raw inputs of previous data in the memory, however, replaying raw pixels is not biologically plausible. Generative replay methods involve training a generative model like an auto-encoder or a generative adversarial network to produce samples from previously learned data \cite {kemker2017fearnet, chenshen2018memory}. However, these approaches are very difficult to train due to issues such as convergence and mode collapse, additionally scaling up generative replay to complex datasets is challenging. Latent replay methods involve storing compressed representations at a specific layer, rather than keeping duplicate copies of input patterns as raw data. These compressed representations capture the essential features of the input data, making them efficient for replay. Utilizing latent replay in LLEDA is not only the most efficient but also a biologically plausible approach \cite{pellegrini2020latent, hayes2020remind, van2020brain}. We summarize LLEDA as follows:
\begin{itemize}
    \item Existing research on combining the lifelong learning and domain adaptation is limited. While some studies like \cite{volpi2021continual} focus on continual and supervised adaptation using labeled data, others such as \cite{wulfmeier2018incremental} and \cite{bobu2018adapting} address continual domain adaptation but assume gradual target shifts, making them less practical.
    \item We present a novel solution called LLEDA, which draws inspiration from the mammalian brain and the CLS theory. LLEDA addresses the issue of catastrophic forgetting and facilitates domain-agnostic knowledge transfer, operating exclusively with unlabeled datasets, allowing learning from a single domain at a time.
    \item LLEDA lies at the intersection of lifelong learning, self-supervised learning, and domain adaptation.
\end{itemize}

\section{Methodology} \label{method}
Our overall objective is to continually update a model to learn distributional shifts while retaining knowledge about past learnings. We propose a novel lifelong domain adaptation framework (depicted in figure \ref{fig-network} and algorithm-\ref{alg:lleda}), which has three key components and is motivated by the CLS theory \cite{mcclelland1995there}. The DA network in LLEDA swiftly adapts to changes in the data distribution between the current domain and previously encountered domains. The SSL network learns to generalise representations through self-supervised learning of domain-agnostic data, while the latent memory component facilitates the interaction between the two networks. By replaying and reactivating past experiences, this component stabilizes new memories for long-term retention and generalization. The combined operation of the DA and SSL networks integrates new information into the long-term network without compromising previous knowledge.

The LLEDA framework process involves the following steps: first, the SSL network learns the visual features and their relationships from the unlabeled input data using self-supervised techniques. As the SSL network is not task-specific, the learned representations are more general, capturing the underlying structure of the data. Next, the DA network uses Maximum Mean Discrepancy (MMD) loss to address domain shift between the current domain and previous domains stored in memory. This loss is backpropagated to both networks for consolidation and to prevent interference. The latent memory component stores and replays past experiences as representations, rather than raw input pixels, to aid interaction between the two networks. All learning occurs in a synchronous and interleaved manner.

\begin{figure*}[t!]
\centering
\hspace*{-1mm}
\includegraphics[width=0.99\textwidth]{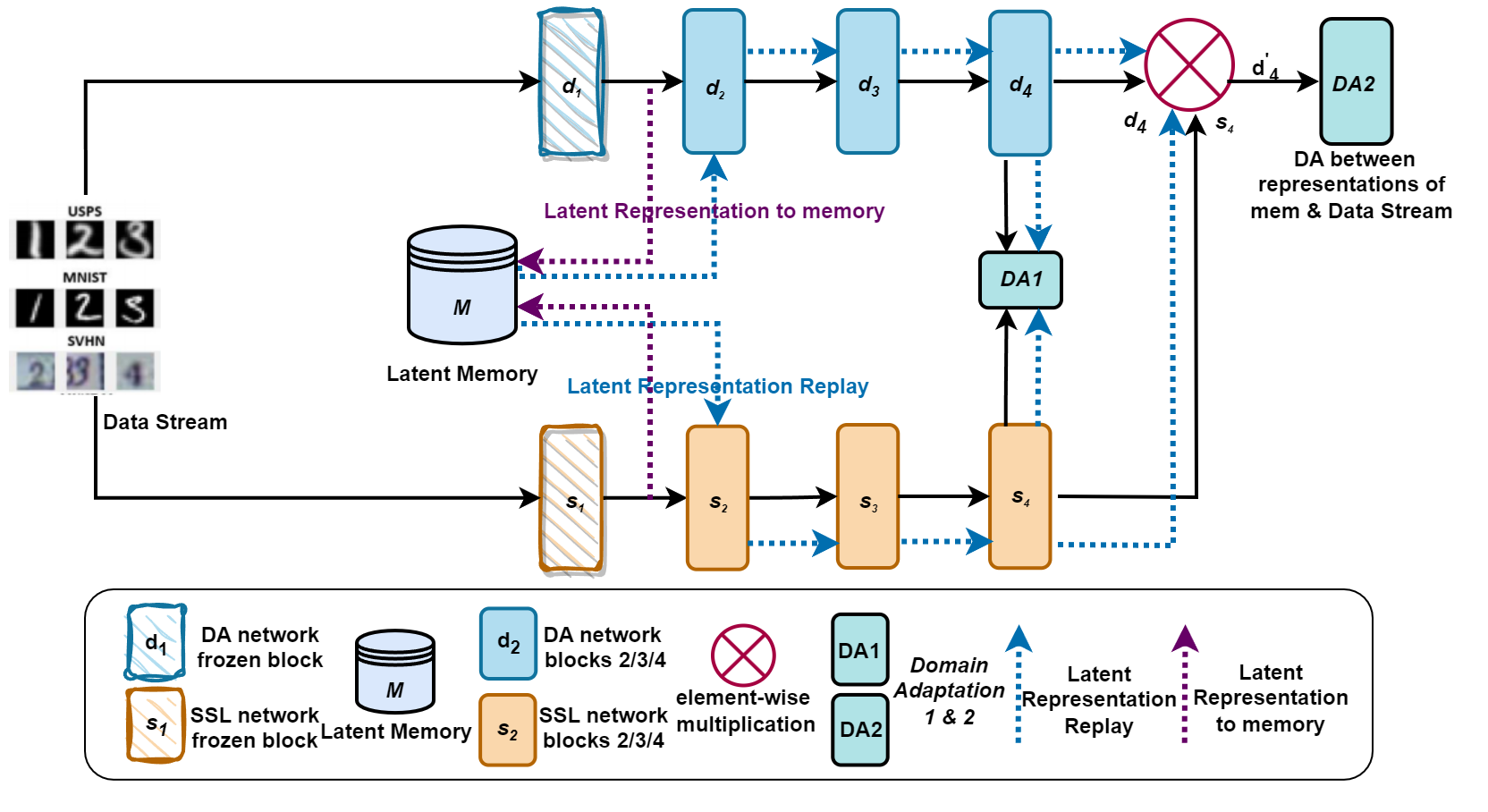}
\vspace*{-0.1cm}
\caption{Overview of latent replay. Demonstration of the flow of latent representations, \textbf{the arrows in blue} show the latent representation flow from memory to the network and \textbf{arrows in pink} show the flow of latent representations from network to memory. }
\label{fig-mem}
\end{figure*}

\subsection{Generalised Feature Learning}
LLEDA employs the SSL network to gradually learn and capture the visual features, underlying structure, and their relationship. As this network is trained independently from the DA network, it does not interfere with the new learnings. Moreover, self-supervised learning provides the model with additional context and information about the input data, enabling it to learn more generic and transferable representations in situations where labeled data is not accessible, which is the case in our scenario. 

LLEDA’s SSL backbone network is compatible with all the existing SSL models (SimCLR \citep{chen2020simple}, BYOL \citep{grill2020bootstrap}, etc.,), so any generic SSL model can be used as the backbone.  However, we have considered VICReg \cite{DBLP:journals/corr/abs-2105-04906} as our backbone to reduce the SSL loss due to its simplicity,  additionally it does not require a memory bank, contrastive samples, or a large batch size.
We have conducted ablation studies using alternative SSL models like SimCLR \citep{chen2020simple} and BYOL \citep{grill2020bootstrap} as our background network to reduce the SSL network's loss, which has been discussed later in section \ref{expabl}. VICReg model uses the weighted average of invariance, variance and covariance to calculate the loss between . The SSL loss is defined as follows:
\begin{equation}
\label{vicreg}
l(z_i, z_j) = \lambda s(z_i, z_j) + \mu[\upsilon(z_i)+ \upsilon(z_j)]+ \nu[c(z_i)+ c(z_j)]
\end{equation}

Where $\lambda$, $\mu$, $\nu$ are the hyper-parameters controlling the importance of each term in the loss.
$s(z_i, z_j)$ is the Invariance,
$c(z_i)$, $c(z_j)$ is covariance and
$\upsilon(z_i)$, $\upsilon(z_j)$ is variance.\\
The overall objective is given by
\begin{equation} 
\label{SSL-loss}
L = \sum_{I \epsilon D} \sum_{t_i, t_j \sim T} l(z_i, z_j)
\end{equation}

\subsection{Domain-specific Representations Learning}
The goal of the DA network is to rapidly learn to reduce the domain discrepancy for the incoming domains, simultaneously working well on the previous domains without catastrophically forgetting the learnings. The DA network uses Maximum Mean Discrepancy (MMD) loss to address domain shift. The DA network, inspired by Dualnet \cite{DBLP:journals/corr/abs-2110-00175}, also interacts with the SSL network and acquires generic representations that influence its learning in a manner akin to biological networks, improving its capacity to reduce discrepancy between domains. It reduces the discrepancy in two stages:
The DA network uses Maximum Mean Discrepancy (MMD) loss to address domain shift. It calculates MMD loss using representations from block 4 of the Resnet (DA1). It again calculates the MMD loss between the memory representations and the current data stream propagation (DA2) following the element-wise multiplication. Calculating the MMD loss at two stages (DA1 and DA2), as seen in figure \ref{fig-mem}, helps to effectively reduce the domain shift, compared to a single domain adaptation loss.
 
Let $s_4$ be the feature representation from the SSL network's residual block, and $d_4$ be the feature representation from the DA network's residual block  as shown in figure \ref{fig-mem}, the adapted feature is obtained during network interaction as follows:
\begin{equation}
\mathrm{d}_{4}^{\prime} = \mathrm{d}_{4} \otimes \mathrm{s}_{4}
\end{equation}

where $\otimes$ denotes the element-wise multiplication,the output of the rapid DA network $d_4$, gradual SSL network $s_4$ and the transformed feature ${d_4}^{\prime}$ all have the same dimension. 

The final layer’s transformed feature ${d_4}^{\prime}$ will be fed into the DA network's head to calculate the DA2 loss using MMD. The rapid DA network takes advantage of the gradual SSL learner’s generalised feature representations resulting in quick adaptation leading to reduced domain shift and improved generalization leading to better identification of classes in the downstream classification task.

MMD defines the distance between the two distributions with their mean embeddings in the Reproducing Kernel Hilbert Space (RKHS). MMD is a two-sample kernel test to determine whether to accept or reject the null hypothesis $p = q$ \cite{gretton2012kernel}, where $p$ and $q$ are source and target domain probability distributions. In short, the MMD between the distributions of two datasets is equivalent to the distance between the sample means in a high-dimensional feature space and is computed by the following equation:

\begin{equation}\label{MMD loss}
L_{{MMD}}  = \left\|\frac{1}{N}\sum_{i=1}^N\phi (x^s_i)-\frac{1}{M}\sum_{j=1}^M\phi (x^t_j) \right\|^2_{H}
\end{equation}

\begin{equation}
\label{MMD loss expand}
\begin{aligned}
 = \frac{1}{N^2}\sum_{i=1}^N\sum_{i^\prime=1}^Nk(x^s_i,x^s_{i^\prime})-\frac{2}{NM}\sum_{i=1}^N\sum_{j=1}^Mk(x^s_i,x^t_j) \\+ \frac{1}{M^2}\sum_{j=1}^M\sum_{j\prime=1}^Mk(x^t_j,x^t_{j^\prime}) 
\end{aligned} \\
\end{equation}
\\
where:\\
$ \phi\left ( . \right ) $ is the mapping to the RKHS H; and \\
$k\left ( .,. \right ) = \left \langle \phi\left ( . \right ), \phi\left ( . \right )  \right \rangle$ is the universal kernel associated with this mapping, and $N, M$ are the total number of items in the source and target respectively. 

\subsection{Latent Replay}
The mammalian brain has successfully evolved to resist catastrophic forgetting by reactivating, replaying, and recreating the experience preserved in memories \cite{ONEILL2010220, wilson1994reactivation}. It retains compressed versions of the crucial information from past experiences and reactivates by replaying these neural activity patterns of prior experiences. Inspired by this, LLEDA stores feature representations from a specific layer instead of raw input pixels. By doing so, it reactivates and replays these representations to overcome catastrophic forgetting. To achieve this, we freeze the layers below the chosen layer, effectively preventing them from being updated during training.

We implement parameter freezing by freezing the layers below block-1, which effectively disables gradient updates for these layers. As a result of this process, the weights in the frozen layers remain unchanged, while the weights in the unfrozen (trainable) layers are allowed to adapt based on the new data. This strategy ensures the preservation of integrity for these layers, promoting stability and accuracy during the replay of stored representations, while also mitigating any potential aging effect \cite{pellegrini2020latent}. Freezing the network also helps with the stability of the stored representations, else they will differ from the feature representations that would have been generated while feed-forwarding from the input layer. In LLEDA, we save the representations from block-1 of our backbone ResNet network into memory and freeze the network layers below block-1 (below the latent replay layer) to prevent them from being updated during the subsequent training on a new task or dataset  and to ensure the stability and accuracy of the representations and to prevent the aging effect.

As our model does not have access to labels, we follow a simple approach of storing a random subset of past latent representations in memory and train the network while interleaving with new domain representations\cite{DBLP:journals/corr/abs-1806-08568}. While selective replay has shown promising results in few settings, several studies have found that random sampling works equally well \cite{hayes2020remind, wu2019large}, achieving similar performance making it a computationally efficient choice, hence we store random subset of representations in the memory. Following that, we save the latent representations from both the DA and the self-supervised networks for the given random image. During memory consolidation, these memories are interleaved with new latent representations to form a more general representation supporting long-term retention and generalization when encountering new domain experiences. To avoid inefficiency, we store only a limited number of latent representations per domain in the memory buffer until it reaches a given number, known as the latent memory size. In our experiments, we tested two sizes: 100 and 250 latent representations. This ensures that the buffer contains a manageable amount of past random experiences at any given time, as depicted in algorithm \ref{alg:memory}.

\begin{algorithm}[H]
    \caption{Pseudocode for the proposed Lifelong Domain Adaptation}
    \label{alg:lleda}
    \SetKwFunction{isOddNumber}{isOddNumber}
    \SetKwInOut{KwInput}{Input}
    \SetKwInOut{KwOutput}{Output}
    \KwInput{Current Domain Data D, Memory M, SSL $\theta$, DA $\phi$}
    \KwOutput{updated $\theta$, $\phi$}
    \For{sampled minibatch ($S_d$, $S_m$) from D and M}
    {     
         Calculate $L_{SSL}$ loss on $S_d$ using  
         equation: \ref{SSL-loss} to update $\theta$ \\
         Calculate $L_{DA1}$ loss on $S_d$ using 
         equation: \ref{MMD loss expand} to update $\phi$ and $\theta$ \\
       
         \If {domain $> 1$} 
            {Calculate $L_{DA1}$ loss on $S_m$ using 
         equation: \ref{MMD loss expand} to update $\phi$ and $\theta$ \\
             Calculate $L_{DA2}$ loss on $S_d$ and  $S_m$ using equation: \ref{MMD loss expand} to update $\phi$ and $\theta$} 
        Add latent representations to memory using algorithm:\ref{alg:memory}\\
    }
\end{algorithm}

\begin{algorithm}[h!]
    \caption{Pseudocode for saving random latent representations to memory}
    \label{alg:memory}
    \SetKwInOut{KwInput}{Input}
    \SetKwInOut{KwOutput}{Output}
    \KwInput{Memory $M$, Representations $R$, Sample Size $s$}
    \KwOutput{Memory $M$}
    $M = \theta$ \\
    $t_m$ = len($M$) \\
    $c_m$ = 0 \\
    \For{each repbatch from $R$}
        {       
                $\delta$ = ${t_m} - {c_m}$  \\
                $h = min(s, \delta)$ \\
                $R_{add}$ = random sampling of size $h$ from repbatch\\
           \eIf {$c_m < t_m$}
                {
                
                $M = M \cup R_{add}$ \\
                $c_m += h$
                }
        {$R_{replace}$ = random sampling of size $s$ from $M$\\
         $M = (M - R_{replace}) \cup R_{add}$}
        }     
\end{algorithm}


       
\section{Experiments \& Results} \label{exp}
\subsection{Datasets} \label{datasets}
We compare and evaluate our method against baseline approaches on a number of benchmark domain adaptation datasets, such as Digits, Office-Home \cite{DBLP:journals/corr/VenkateswaraECP17}, Office-CalTech \cite{gong2012geodesic} and ImageCLEF-DA.

\begin{figure}[h!]
\centering
\includegraphics[width=0.9\linewidth]{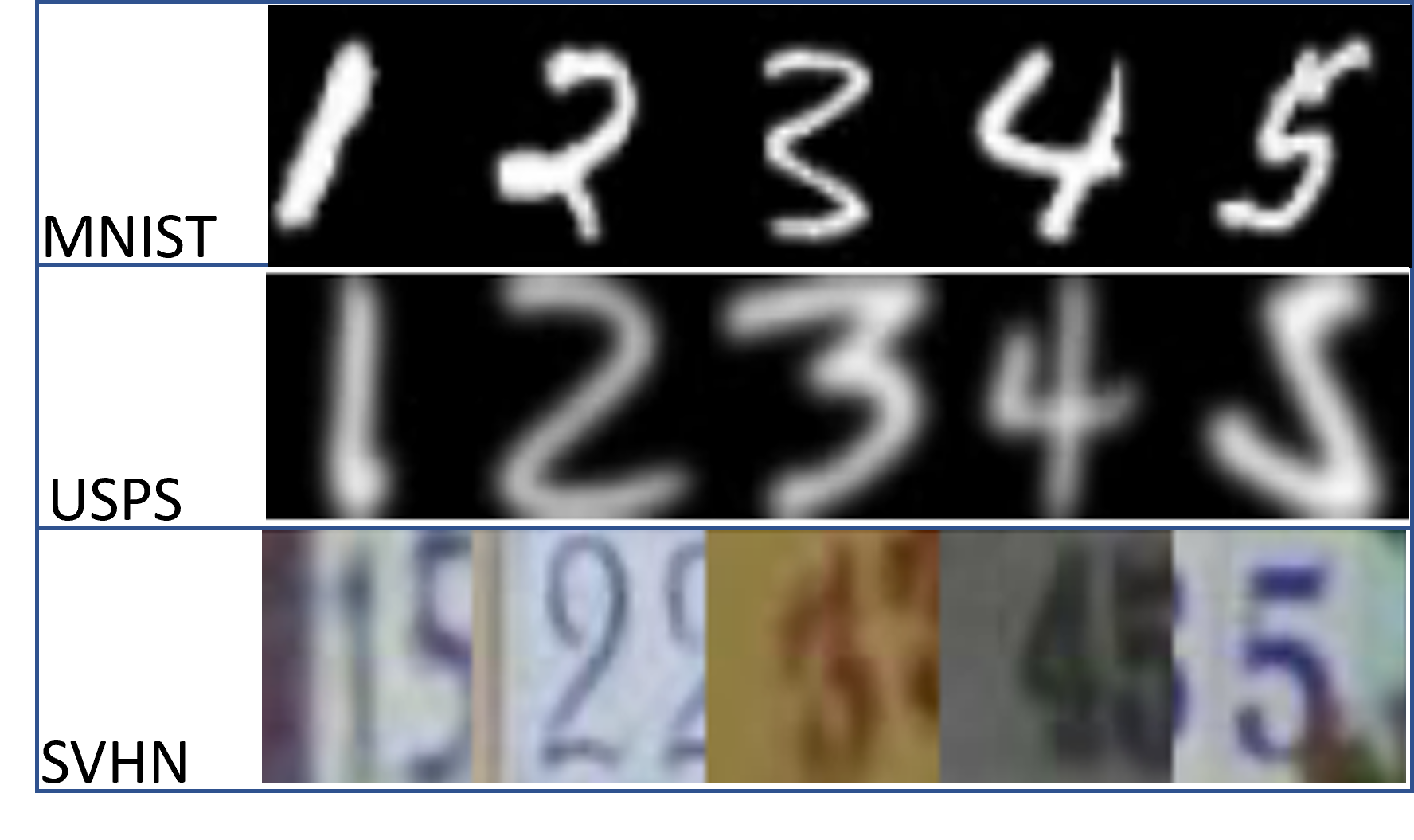}
\caption{Sample images from Digits Dataset.}
\label{fig-digits}
\end{figure}

\textbf{Digit Dataset:} We consider the standard digits dataset broadly adopted by the computer vision community. MNIST \cite{lecun1998gradient} and USPS \cite{denker1989neural} are hand-written grey-scale images, with relatively small domain differences. SVHN \cite{netzer2011reading} contains images of street numbers with more than one digit in each image. We conducted experiments on two tasks: SVHN → USPS → MNIST and MNIST → USPS → SVHN and reported the average accuracy of the trained model in the context of lifelong learning setting. These two scenarios will allow us to reflect on the performance of lifelong learning scenarios starting from easy datasets, moving to harder ones and vice versa. Sample images of the digit dataset are presented in figure \ref{fig-digits}. 

\begin{figure}[h!]
\centering
\hspace*{-1mm}
\includegraphics[width=0.9\linewidth]{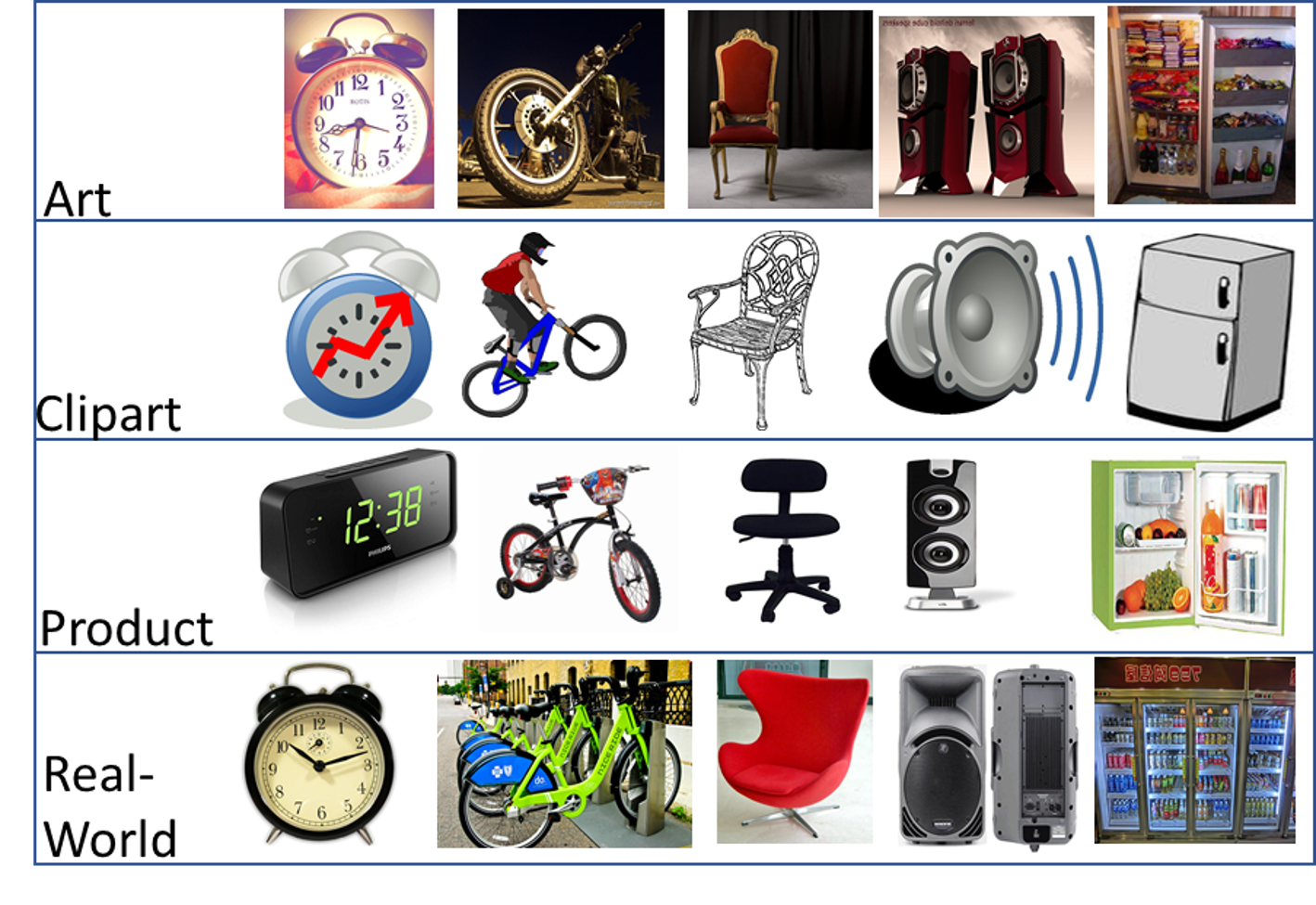}
\vspace*{-0.1cm}
\caption{Sample Images from Office-Home Dataset.}
\label{fig-office-home}
\end{figure}

\textbf{Office-Home \cite{DBLP:journals/corr/VenkateswaraECP17}:} The office-home data consists of four visual domains: Art (A), Clipart (C), Real World (R), and Product (P) each consisting of images from 65 visual categories totalling 15,500 images in office and home settings leading to the possibility of defining 12 pair-wise binary UDA tasks. We conducted several experiments on two tasks: Art → Realworld → Clipart → Product and Product → Clipart → Realworld → Art and reported the average accuracy of the trained model in the context of lifelong learning setting. Sample images of the office-home dataset are presented in figure \ref{fig-office-home}.

\begin{figure}[h!]
\centering
\includegraphics[width=0.9\linewidth]{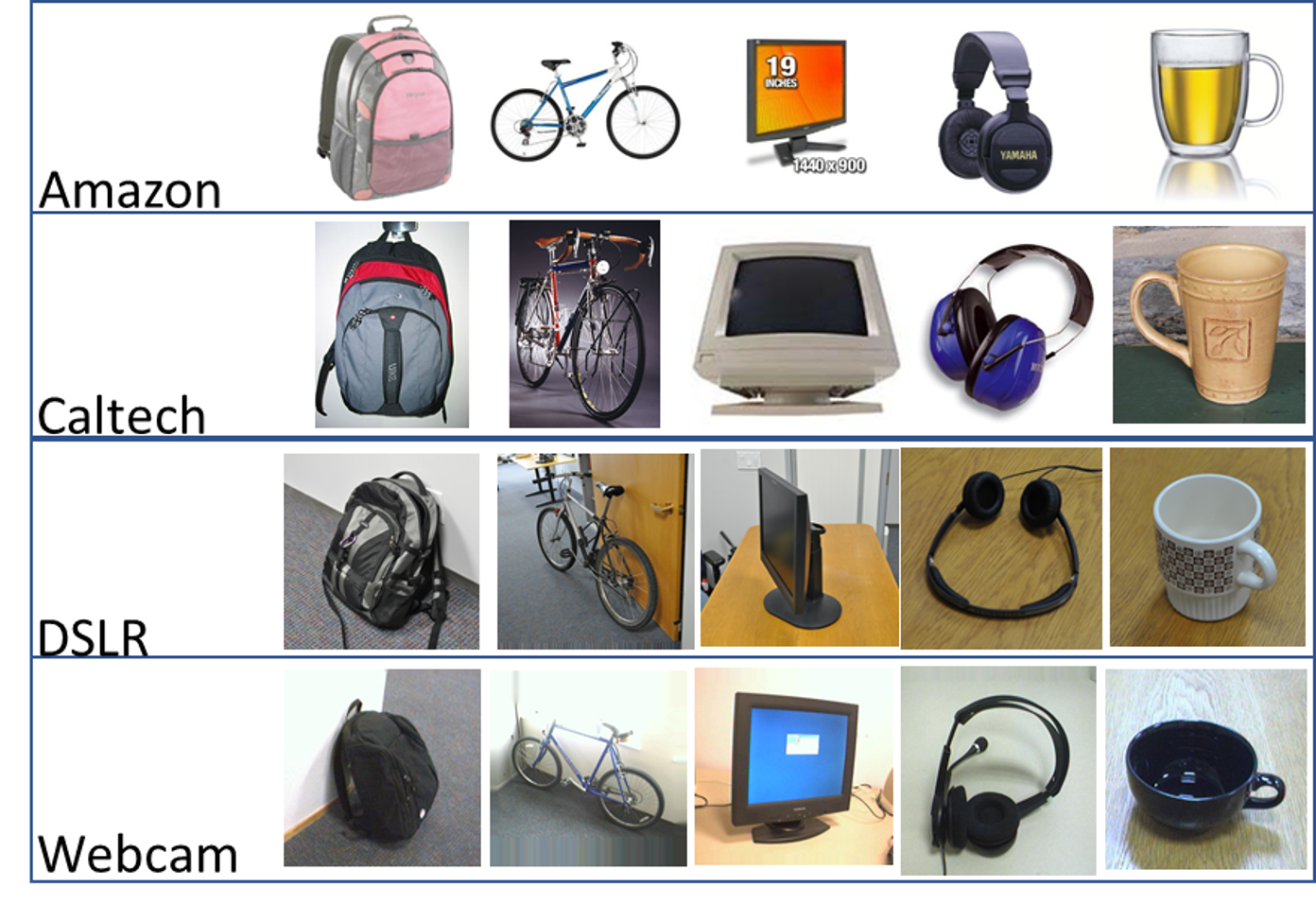}
\caption{Sample Images from Office-Caltech Dataset.}
\label{fig-office-caltech}
\end{figure}

\textbf{Office-CalTech \cite{gong2012geodesic}:} This dataset is an extension of the Office-31 \cite{Saenko2010AdaptingVC} with 10 common categories shared by Office-31 and the CalTech-256 dataset \cite{griffin2007caltech}. This dataset has four domains: Webcam (W), DSLR (D), Amazon (A), and CalTech (C). We conducted several experiments on two tasks: DSLR → Webcam → Amazon → Caltech  and Caltech → Amazon → Webcam → DSLR and reported the average accuracy of the trained model in the context of lifelong learning setting. Sample images of the office-caltech datasets are presented in figure \ref{fig-office-caltech}.

\begin{figure}[h!]
\centering
\includegraphics[width=0.875\linewidth]{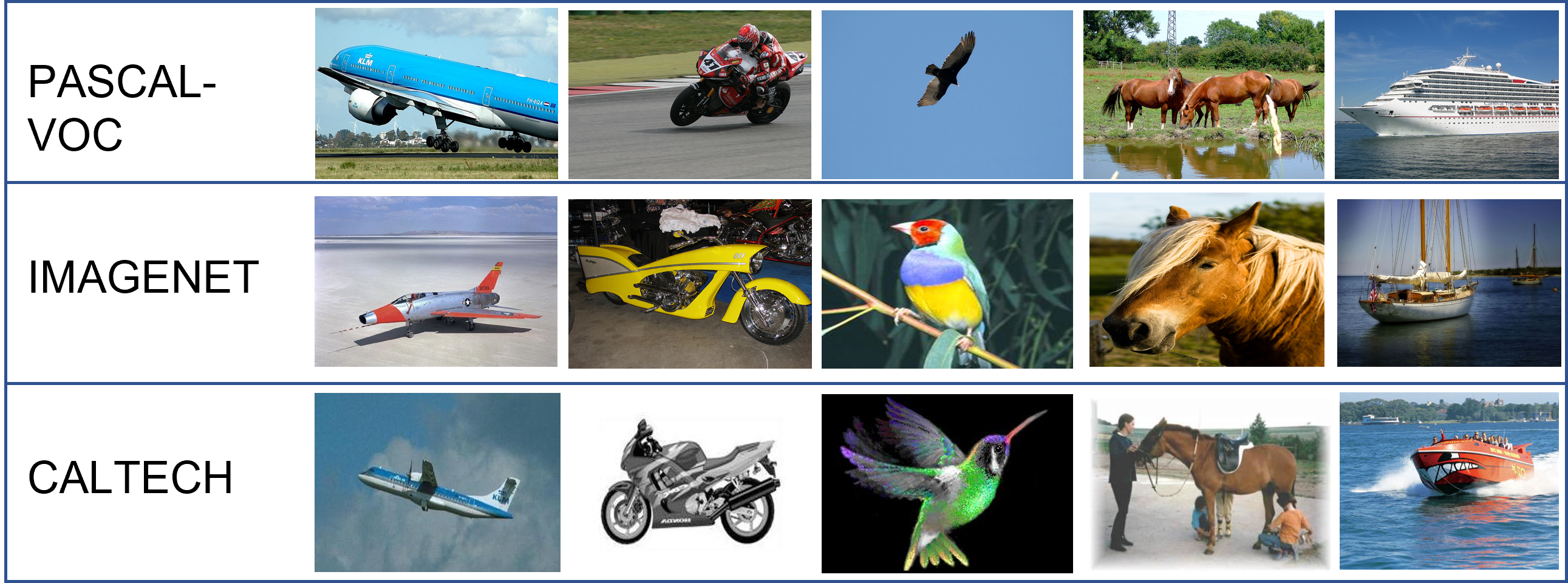}
\caption{Sample images from ImageCLEF-DA Dataset.}
\label{fig-ImageCLEF}
\end{figure}

\textbf{ImageCLEF-DA:} This dataset has four domains with twelve categories each: Caltech-256, ImageNet ILSVRC 2012, and Pascal VOC 2012. We conducted several experiments on two tasks: Caltech → ImageNet → Pascal and Pascal → ImageNet → Caltech and reported the average accuracy of the trained model in the context of lifelong learning setting. Sample images of the ImageCLEF dataset are presented in figure-\ref{fig-ImageCLEF}.

\subsection{Training Methods}
We benchmark LLEDA  against the baseline method which uses a single network and finetunes the model as the new training domains come along, we then compare our LLEDA methodology with DANN \cite{ganin2016domain} and DAN \cite{long2015learning}, both of them are classic domain adaptation methods and both these methods have access to source and target data during training. We also compare LLEDA with CUA \cite{bobu2018adapting} and GRCL \cite{tang2021gradient} which are continual learning replay-based methods. It is important to note that both methods have access to source-labeled data, unlike LLEDA which operates without labeled data from either the source or target domains. We made an exciting observation during experimentation by increasing the size of latent representations stored in memory from 100 to 250, resulting in impressive results. Consequently, we tested LLEDA-100 and LLEDA-250, with 100 and 250 latent representations stored in the memory respectively. It's worth mentioning that other papers typically use a memory size of 2000 images/representations. Furthermore, we compare these methods with the supervised version of our approach, LLEDA-S. Most of the methods provide the results in the domain adaptation setting, but we have provided our results in the context of lifelong learning setting. 

\subsection{Implementation Details}
Our implementation involves three stages. In the first stage, we pre-train the model on ImageNet which serves as the foundation for subsequent stages, which we call as a pre-trained model. In the second stage, we use the pre-trained model and further train the LLEDA model as outlined in the methodology section. 
In the final stage, we freeze the trained network and train a linear classifier on top of the fixed representation, while removing the MMD projection head. This trained linear classifier is used for evaluation purposes. 

For the pretraining phase, we employ the ResNet18 \citep{he2016deep} architecture as our backbone model, pretrained on the ImageNet dataset. During this phase, we use two nodes, each equipped with 4 V100 GPUs. The training process is carried out using the LARS optimizer \cite{you2017large} with a batch size of 512, and we apply a weight decay of le-6 training for a total of 100 epochs. During the subsequent training phase, we use the pretrained network obtained from the previous stage as a starting point. We incorporate the stored latent representations from the layer-1 and combine it with the current domain data representations. Finally, during finetuning phase, we freeze the trained network and further train a linear classifier on top of this fixed representations whilst discarding the MMD part of the network. We use the resulting network to evaluate on the domain datasets to assess its performance.
Similar to most self-supervised models \cite{chen2020simple, chen2020big, oord2018representation, chen2020big, he2020momentum, bardes2021vicreg, zbontar2021barlow, grill2020bootstrap, chen2021exploring}, we report performance by training a linear classifier on top of a fixed representation to evaluate representations which is a standard benchmark that has been adopted by many papers in the literature.




\subsection{Results and Analysis}
Our primary objective is to evaluate the performance of our proposed LLEDA framework in lifelong learning domain adaptation scenarios. This assessment involves sequentially training the model on different domains. This sequential training process, which we refer to as a "cycle," involves training the model on one domain, followed by training it on the next domain, and so on. Upon the completion of each cycle, we consider the resulting model as the "final model". This final model is then tested on all the domains it was trained on, and the corresponding results are presented. Additionally, we calculate the average performance of the final model across each domain within every cycle. It is crucial to note that the other state-of-the-art methods, which we refer to, operate based on the UDA evaluation criteria. These methods are trained on a labeled source dataset and tested on an unlabeled target dataset, but they are not continually trained as in our case. Therefore, our results reflect the challenging scenario where testing is performed within cycles, and without access to any labeled data. 
\begin{table}[]
\centering
\begin{tabular}{lll}
\hline 
Dataset                 & Method   & Avg \\ \hline \hline
\multirow{7}{*}{Digits} & Baseline & 56.7        \\
                        & DANN     & 74.5    \\
                        & DAN      & 72.9    \\
                        & CUA      & 82.1        \\
                        & GRCL     & 85.3    \\
                        & LLEDA-S  & 89.0     \\
                        & LLEDA-100   & 86.6   \\
                        & \textbf{LEDA-250}    & \textbf{89.5}   \\ \hline
\end{tabular}
\caption{ Comparison of the proposed LLEDA method on Digit datasets comprising MNIST, USPS and SVHN domains with state-of-the-art methods, using Average Accuracy (Avg) across the domains as the performance metric. LLEDA-100 and 250 represent the latent memory size of 100 and 250. LLEDA-S is a supervised model with access to labels. The best average is indicated in \textbf{bold}.}
\label{tab:my-Digits}
\end{table}

\begin{table}[]
\centering
\begin{tabular}{lll}
\hline 
Dataset                 & Method   & Average \\ \hline \hline
\multirow{6}{*}{Office-Home} & Baseline &   28.7      \\
                        & DANN     & 57.6    \\
                        & DAN      & 56.3    \\
                        & CUA      &  58.6    \\
                        & LLEDA-S  & 60.3      \\
                         & LLEDA-100   & 58.2   \\
                        & \textbf{LLEDA-250}  & \textbf{62.1}  \\ \hline
\end{tabular}
\caption{ Comparison of the proposed LLEDA method on Office-Home datasets comprising Art, Clipart, Product and Real-World datasets with state-of-the-art methods, using Average Accuracy (Avg) across the domains as the performance metric. LLEDA-100 and 250 represent the latent memory size of 100 and 250. LLEDA-S is a supervised model with access to labels. The best average is indicated in \textbf{bold}.}
\label{tab:my-O-H}
\end{table}

\begin{table}[]
\centering
\begin{tabular}{lll}
\hline 
Dataset                 & Method   & Average \\ \hline \hline
\multirow{7}{*}{Office-Caltech} & Baseline &   52.3     \\
                        & DANN     & 81.7    \\
                        & EWC      & 84.5    \\
                        & CUA      & 84.8    \\
                        & GRCL     & 87.2    \\
                        & LLEDA-S  & 87.5     \\
                        & LLEDA-100    & 86.1   \\ 
                         & \textbf{LLEDA-250}    & \textbf{90.3}   \\ \hline
\end{tabular}
\caption{Comparison of the proposed LLEDA method on Office-Caltech datasets comprising Amazon, Caltech, DSLR and webcam domains with state-of-the-art methods, using Average Accuracy (Avg) across the domains as the performance metric. LLEDA-100 and 250 represent the latent memory size of 100 and 250. LLEDA-S is a supervised model with access to labels. The best average is indicated in \textbf{bold}.}
\label{tab:my-O-C}
\end{table}

\begin{table}[]
\centering
\begin{tabular}{lll}
\hline 
Dataset                 & Method   & Average \\ \hline \hline
\multirow{6}{*}{ImageCLEF} & Baseline &    51.3     \\
                        & DANN     &  82.2 \\
                        & DAN      &   82.4  \\
                        & LLEDA-S &       90.3 \\
                         & LLEDA-100   &   89.7 \\
                        & \textbf{LLEDA-250} &   \textbf{92.6}   \\ \hline
\end{tabular}
\caption{ Comparison of the proposed LLEDA method on ImageCLEF-DA datasets comprising Caltech, ImageNet ILSVRC, and Pascal-VOC domains with state-of-the-art methods, using Average Accuracy (Avg) across the domains as the performance metric. LLEDA-100 and 250 represent the latent memory size of 100 and 250. LLEDA-S is a supervised model with access to labels. The best average is indicated in \textbf{bold}.}
\label{tab:my-ImageCLEF}
\end{table}

\begin{table*}
\centering
\begin{tabular}{@{}lllllllll@{}}

\toprule
\multirow{2}{*}{Method} &
  \multicolumn{3}{c}{CYCLE-1} &
   &
  \multicolumn{3}{c}{CYCLE-2} &
  \multirow{2}{*}{Avg} \\ \cmidrule(lr){2-8}
 &
  \multicolumn{1}{c}{SVHN} &
  \multicolumn{1}{c}{USPS} &
  \multicolumn{1}{c}{MNIST} &
   &
  \multicolumn{1}{c}{MNIST} &
  \multicolumn{1}{c}{USPS} &
  \multicolumn{1}{c}{SVHN} &
   \\ \midrule \hline
LLEDA-VICReg & 71.3 & 93.3 & 94.1 &  & 86.7 & 85.9 & 88.7 & 86.6 \\ 
LLEDA-SimCLR   & 73.6 & 94.8 & 93.8 &  & 78.9 & 87.2  & 90.5 & 86.4 \\
LLEDA-BYOL & 70.9 & 95.5 & 92.6 &  & 86.3 & 88.9 & 87.5 & 86.9 \\ \bottomrule
\end{tabular}
\caption{Comparison of the proposed LLEDA's SSL network  using state-of-the-art self-supervised methods as building blocks.}
\label{tab:lleda-ssl}
\end{table*}

\begin{table}[]
\centering
\begin{tabular}{lll}
\hline 
Dataset                 & Method   & Average \\ \hline \hline
\multirow{4}{*}{Digits} & Elementwise multiplication &   \textbf{86.6}    \\
                        & Elementwise addition     &    75.3 \\
                        & Elementwise maximum      &    39.3 \\
                        & Elementwise mean      &     71.9 \\
                        \hline
\end{tabular}
\caption{Comparison of LLEDA's DA and SSL network interaction using various element-wise operations. The best average is indicated in \textbf{bold}.}
\label{tab:cons}
\end{table}
Baseline: Initially, we train a basic model, denoted as $M_i$, on the domain $D_i$. As new domains become available, we fine-tune the model by training it on the subsequent domain, $D_{i+1}$. However, we observe that at the end of the cycle, this approach tends to exhibit poor performance on earlier domains due to a phenomenon known as CF. This outcome serves as our baseline for comparison in this study. In our experiments, we use Resnet18 as the baseline model and to assess the effectiveness of our proposed method, we evaluate its performance against the baseline.

\textbf{Digits dataset}: The table-\ref{tab:my-Digits}  presented in this study showcases the average performance of different methods on the Digits dataset, encompassing MNIST, USPS, and SVHN domains in the lifelong learning scenario cycle. The proposed method, LLEDA, exhibits a remarkable advantage over other approaches by effectively handling sequential training in lifelong learning adaptation scenarios. LLEDA achieves an impressive average accuracy of 89.5\%, surpassing the baseline accuracy of 56.7\%, and outperforms several state-of-the-art methods, including DANN (74.5\%), DAN (72.9\%), CUA (82.1\%), and GRCL (85.3\%). These results underscore the superior performance of LLEDA in the context of lifelong learning, highlighting its adaptability and sequential learning capabilities. It is worth noting that LLEDA-S demonstrates improved performance compared to LLEDA-100, which can be attributed to the availability of labeled data. Additionally, the performance gap between LLEDA-100 and LLEDA-250 can be linked to the amount of additional memory representations saved. With more representations stored and replayed, LLEDA-250 accumulates a diverse set of samples from various domains or tasks, enabling the model to develop more robust and generalizable features, resulting in enhanced performance on new data.

\textbf{Office-Home dataset}: The table-\ref{tab:my-O-H} presents the average performance of different methods on the Office-Home dataset involving Art, Clipart, Product and Real-world domains, and focuses on the lifelong domain adaptation cycle. The baseline method achieves an average accuracy of 28.7\%, which is relatively low. In contrast, our proposed LLEDA method without access to labels, achieves a substantial increase in average accuracy, rising from 28.7\% to an impressive 62.1\%. Similar to the findings in the digits dataset, LLEDA-S outperforms LLEDA-100 as expected. Moreover, LLEDA-250 achieves the highest average accuracy among the other state-of-the-art methods suggesting its superior performance effectively handling the lifelong learning adaptation scenarios on the Office-Home dataset.

\textbf{Office-Caltech dataset}: The table-\ref{tab:my-O-C} presents the average accuracy of various methods on the Office-Caltech dataset, along with Amazon, Caltech, DSLR and Webcam domains. It summarises the average performance of different methods on the Office-Caltech dataset. The baseline method achieves an average accuracy of 52.3\%, while DANN performs significantly better at 81.7\%. EWC and CUA further improve the results with average accuracies of 84.5\% and 84.8\%, respectively. GRCL shows even higher performance with 87.2\% accuracy, and LLEDA-S follows closely with 87.5\% accuracy. LLEDA-100 achieves an average accuracy of 86.1\%, and the top-performing method in this dataset is LLEDA-250, achieving an impressive average accuracy of 90.3\% without utilising any labeled data. These results demonstrate the superiority of LLEDA in effectively addressing lifelong learning adaptation scenarios within the Office-Caltech dataset, outperforming other state-of-the-art methods and indicating its potential as a robust approach for continual domain adaptation tasks.


\textbf{ImageCLEF-DA dataset}: The table-\ref{tab:my-ImageCLEF} presents the average accuracy of various methods on the ImageCLEF dataset, along with Pascal VOC, ImageNet, and Caltech domains. It summarises the average performance of different methods on the ImageCLEF-DA dataset. The results indicate that the ImageCLEF Baseline method achieved the lowest average accuracy of 51.3\%, suggesting limited performance in handling domain shifts. DANN and DAN, which employ domain adaptation techniques, showed significant improvement with average accuracies of 82.2\% and 82.4\%, respectively. However, their performance was surpassed by the LLEDA-250 method, which achieved an impressive average accuracy of 92.6\%. Both LLEDA-100 and LLEDA-250 outperformed DANN and DAN, with average accuracies of 89.7\% and 92.6\%, respectively. Similar to the domains above LLEDA-S has an increased performance in comparison to LLEDA-100 due to the access of labeled data. LLEDA-250 emerged as the top-performing method across all datasets, suggesting its superiority in addressing continual domain adaptation tasks effectively.

The experimental results conducted on the Digits, Office-Home, Office-Caltech and ImageCLEF-DA datasets consistently demonstrate the superior performance of the LLEDA framework over other state-of-the-art methods in addressing lifelong learning scenarios. LLEDA effectively tackles challenges such as catastrophic forgetting and domain shift through sequential access to unlabeled data, showcasing its adaptability.  Despite variations in evaluation setups, LLEDA showcases impressive performance in handling domain shifts even without label access, resulting in consistently high average accuracy. These results affirm the robustness and potential of the LLEDA framework in addressing lifelong learning challenges across diverse datasets.

\subsection{Ablation Studies} \label{expabl}
\textbf{Ablation: LLEDA's SSL network using state-of-the-art self-supervised methods as building blocks}
We evaluated the effectiveness of LLEDA by replacing the LLEDA's SSL network with some of the state-of-the-art SSL networks. Our objective is to assess the LLEDA's performance in lifelong learning scenarios by sequentially training on different domains. To assess lifelong learning performance, we start by training the image samples from one domain, followed by training on the next domain, and so on. We call this sequential training process as a cycle. For example, in cycle-1 (SVHN - USPS - MNIST), we trained the LLEDA model on the SVHN, followed by training on the USPS, and finally on the MNIST. Similarly, in cycle-2 (MNIST - USPS - SVHN), we trained on MNIST followed by training on USPS, and finally training on SVHN. Each cycle represents a sequential training process on different datasets.


We analysed the accuracy of LLEDA to investigate the impact of the SSL network selection on gradual learning network.  In table \ref{tab:lleda-ssl}, we compare three SSL methods- SimCLR \citep{chen2020simple}, BYOL \citep{grill2020bootstrap} and VICReg. We chose these SSL networks as all three methods feature different losses and use different techniques to avoid collapse such as negative samples, redundancy reduction, etc. Additionally, the former is a contrastive-based method, whereas the latter two are non-contrastive ones. 

Table-\ref{tab:lleda-ssl} shows that the average performance of VICReg is robust in comparison to the average performance of contrastive-based SimCLR \citep{chen2020simple} as the latter requires large amounts of contrastive pairs and a higher batch size to converge. The average performance of VICReg slightly underperforms compared to BYOL \citep{grill2020bootstrap}. Overall, the comparative performance of all three SSL methods with respect to the LLEDA framework is almost relatively similar, with minor variations in accuracy across different datasets and cycles. This similarity can be attributed to the fact that SSL is not directly employed for the downstream task in LLEDA. Instead, element-wise multiplication balances the contributions of both SSL and DA networks. This network integration into the downstream tasks may explain the comparable performance across various SSL techniques. Consequently, these findings suggest that LLEDA's gradual learning network can effectively accommodate the substitution of any generic SSL method, ensuring its efficiency and adaptability to different SSL approaches. 

\textbf{LLEDA SSL and DA network interaction using element-wise operations:}
We analysed different types of operations used for interactions and influence between the DA network and the SSL network. We considered element-wise addition, element-wise maximum value, and element-wise mean besides adapting element-wise multiplication to test the generalization ability. Table \ref{tab:cons} demonstrates that the element-wise maximum value seems like a poor choice since the interaction between the two networks appears more competitive than complementary. Element-wise multiplication excels at emphasizing agreement between networks, resulting in a combined representation that is more domain-invariant and robust. In contrast, element-wise addition and element-wise show some promise compared to element-wise maximum value, but fall short in capturing complementary features and striking an optimal balance between SSL and DA networks, ultimately leading to subpar model performance. 
Therefore, the adapted element-wise multiplication emerges as the most favorable choice offering superior generalization capabilities.


\section{Conclusion \& Future Work} \label{con-fw}
Inspired by how the human brain works and the CLS theory, we developed LLEDA, a model that can perform competitively in a lifelong domain adaptation setting across several standard benchmark datasets. Our experiments demonstrate that LLEDA can effectively tackle downstream domain adaptation tasks without access to labeled data, outperforming several other existing methods. This is a very exciting line of work, as in many real-life settings, e.g. healthcare and agriculture, one does not have the luxury of large, curated and labeled data. With methods like LLEDA, we can leverage such datasets to learn continuously and improve performance. 

We believe our work will encourage future research in lifelong domain adaptation using unlabeled source and target domain data, as this is a more realistic scenario in several real-life settings. As our next step, we aim to investigate efficient lossy and lossless compression techniques for compressing latent representations in LLEDA, as well as show how LLEDA performs in larger datasets, which require extensive computational resources. Another line of work will involve exploring techniques, e.g. through distillation and quantization, that will reduce the computational overhead required.

\section*{Credit authorship contribution statement}

\textbf{Mamatha Thota:} Conceptualization, Data Curation, Investigation, Methodology, Software, Formal Analysis, Writing – Original Draft. \textbf{Dewei Yi:} Methodology, Supervision, Writing – Review \& Editing. \textbf{Georgios Leontidis:} Conceptualization, Investigation, Methodology, Validation, Project Administration, Supervision, Writing – Review \& Editing.

\section*{Declaration of Competing Interest}
The authors declare that they have no known competing financial interests or personal relationships that could have appeared to influence the work reported in this paper.

\section*{Acknowledgments}
This work used the Cirrus UK National Tier-2 HPC Service at EPCC (http://www.cirrus.ac.uk). Access granted through the project: ec173 - Next-gen self-supervised learning systems for vision tasks.

\end{document}